\documentclass{article}

\usepackage[nonatbib,final]{neurips_2021}
\usepackage[utf8]{inputenc} % allow utf-8 input
\usepackage[T1]{fontenc}    % use 8-bit T1 fonts
\usepackage{hyperref}       % hyperlinks
\usepackage{url}            % simple URL typesetting
\usepackage{booktabs}       % professional-quality tables
\usepackage{amsfonts}       % blackboard math symbols
\usepackage{nicefrac}       % compact symbols for 1/2, etc.
\usepackage{microtype}      % microtypography
\usepackage{xcolor}         % colors

\usepackage{amsmath,amssymb,amsthm}
\usepackage[ruled,vlined,linesnumbered]{algorithm2e}
\usepackage{verbatim}
\usepackage{braket}
\usepackage{todonotes}
\usepackage{paralist}
\usepackage{caption}
\usepackage{bookmark}
\usepackage{subfigure}
\usepackage{titlesec}
\usepackage{appendix}
\usepackage{authblk}

\makeatletter
\newcommand{\printfnsymbol}[1]{%
  \textsuperscript{\@fnsymbol{#1}}%
}
\makeatother

\definecolor{myorange}{RGB}{245,156,74}
\hypersetup{
	colorlinks=true,
	linkcolor=red,
	filecolor=blue,      
	urlcolor=red,
	citecolor=-myorange,
}

\usepackage{floatrow}

\newfloatcommand{capbtabbox}{table}[][\FBwidth]

\title{What Makes Multi-modal Learning\\ Better than Single (Provably)}
\newtheorem{theorem}{Theorem}

\newtheorem{assumption}{Assumption}
\newtheorem{proposition}{Proposition}
\newenvironment{proofof}[1]{\begin{proof}[Proof of {#1}]}{\end{proof}}
\newtheorem{lemma}{Lemma}

\newenvironment{definition}[2][Definition]{\begin{trivlist}
\item[\hskip \labelsep {\bfseries #1}\hskip \labelsep {\bfseries #2.}]}{\end{trivlist}}

\author{
Yu Huang$^{1,}$\thanks{equal contribution}~, Chenzhuang Du$^{1,}$\printfnsymbol{1}, Zihui Xue$^{2}$, Xuanyao Chen$^{3,4}$,\and  Hang Zhao$^{1,4}$, Longbo Huang$^{1,}$\thanks{Correspondence to: longbohuang@tsinghua.edu.cn}\\
\small{$^1$ Institute for Interdisciplinary Information Sciences, Tsinghua University}\\
\small{$^2$ The University of Texas at Austin}\quad \small{$^3$ Fudan University}\\
\small{$^4$ Shanghai Qi Zhi Institute}
}

\begin{document}
\maketitle

\begin{abstract}
%Human perception of the world is inherently multi-modal.
The world provides us with data of multiple modalities.
Intuitively, models fusing data from different modalities outperform their uni-modal counterparts, since more information is aggregated. Recently, joining the success of deep learning, there is an influential line of work on deep multi-modal learning, which has remarkable empirical results on various applications. However, theoretical justifications in this field are notably lacking.
\begin{center}
    \textit{Can multi-modal learning provably perform better than uni-modal?} 
\end{center}
In this paper, we answer this question under a most popular multi-modal fusion framework, which firstly encodes features from different modalities into a common latent space and seamlessly maps the latent representations into the task space. 
We prove that learning with multiple modalities achieves a  smaller population risk than only using its subset of modalities. The main intuition is that the former has a more accurate estimate of the latent space representation. To the best of our knowledge, this is the first theoretical treatment to capture important qualitative phenomena observed in real
multi-modal applications from the generalization perspective. Combining with experiment results, we show that multi-modal learning does possess an appealing formal guarantee.
\end{abstract}

\section{Introduction}

% The world provides us with multi-modal data, allowing us to build multi-sensor systems to perceive it. As deep learning develops, more and more multi-modal applications have sprung up like mushrooms after a spring rain. For example, PixelPlayer~\cite{zhao2018sound} learns to find the sound source from the image and separate the sounds into different components which can represent the sound from each pixel. \cite{liang2018deep} designs a reliable and efficient end-to-end learnable 3D object detector to perform accurate localization by exploiting both LIDAR as well as cameras.  CLIP~\cite{radford2021learning} demonstrates that simple pre-training task between text and image is an efficient way to learn SOTA image representations and benchmarks over 30 different vision datasets.
%It is encouraging, however, to our best knowledge, deep multi-modality learning lacks theoretical guarantee.
Our perception of the world is based on different modalities, \textit{e.g.} sight, sound, movement, touch, and even smell~\cite{smith2005development}. Inspired from the success of deep learning~\cite{krizhevsky2012imagenet, he2016deep}, deep multi-modal research is also activated, which covers fields like audio-visual learning~\cite{chen2020vggsound, wang2020makes}, RGB-D semantic segmentation~\cite{seichter2020efficient,hazirbas2016fusenet} and Visual Question Answering~\cite{goyal2017making, anderson2018vision}.

While deep multi-modal learning shows excellent power in practice, theoretical understanding of deep multi-modal learning is limited.
Some recent works have been done towards this direction~\cite{sun2020tcgm,zhang2019cpm}. However, these works made strict assumptions on the probability distributions across different modalities, which may not hold in real-life applications~\cite{ngiam2011multimodal}. Notably, they do not take \textit{generalization} performance of multi-modal learning into consideration. %, since the relationships across different modalities are usually uncertain in practice~\cite{ngiam2011multi-modal}.}%Deep multi-modal learning shows great power in practice. However, from a theoretical standpoint, our understanding towards deep multi-modal learning is extremely limited. Recently, there has been a research aiming to obtain the principled understanding of multi-modal learning in theory. \cite{sun2020tcgm,zhang2019cpm} provides theoretical guarantee for an information theory based approach proposed for semi-supervised multi-modal learning. Besides, while very little is known about the theory of multi-modal learning, there is a close
%cousin of the setting, called multi-view learning~\cite{xu2013survey}, which already has comprehensive theoretical underpinnings. Yet, two common principles, \textit{consensus} and \textit{complement}, which are crucial in the theoretical analysis  of multi-view learning, are not applicable in multi-modal learning. 
Toward this end, the following fundamental problem remains largely open:

\begin{center}
    \textit{Can multi-modal learning provably performs better than uni-modal?}
\end{center} 

In this paper, we provably answer this question from two perspectives:
\begin{itemize}
    \item (When) Under what conditions multi-modal performs better than uni-modal?
    \item (Why) What results in the performance gains ?
\end{itemize}

The framework we study is abstracted from the multi-modal fusion approaches, which is one of the most researched topics of multi-modal learning~\cite{baltruvsaitis2018multimodal}. %We study a widely used composite multi-modal framework~\cite{zhou2020multi}, which can seamlessly perform latent space learning and task-specific learning in a unified framework.
Specifically, we first encode the complex data from heterogeneous sources into a common latent space $\mathcal{Z}$. The true latent representation is $g^{\star}$ in a function class $\mathcal{G}$, and the task 
mapping $h^{\star}$ is contained in a function class $\mathcal{H}$ defined on the latent space. Our model corresponds to the recent progress of deep multi-modal learning on various applications, such as scene classification~\cite{du2021improving} and action recognition~\cite{kalfaoglu2020late, wang2020makes}.

\begin{figure}[t]
\includegraphics[width=7cm,height=4cm]{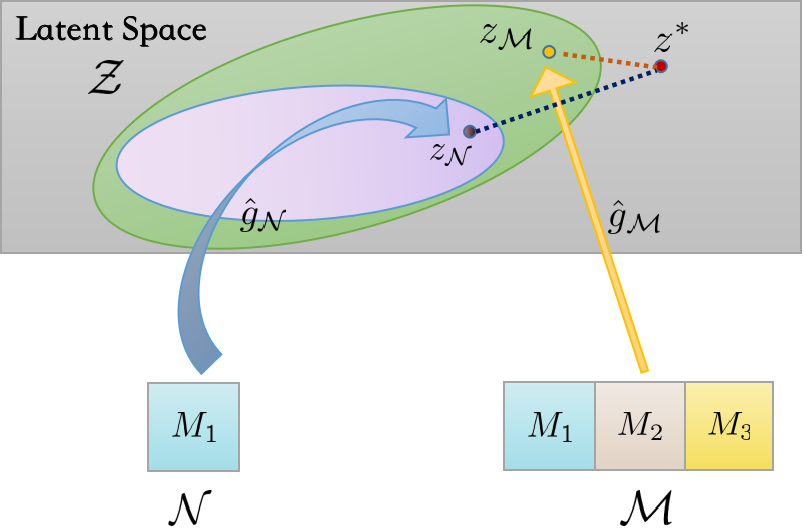}
\centering
\caption{\textbf{\textit{$\mathcal{M}$ vs $\mathcal{N}$} modalities latent space representation}, where the latter is a subset of the former. $z_{\mathcal{M}}$, $z_{\mathcal{N}}$ and $z^{\star}$ are  images on the latent space $\mathcal{Z}$ corresponding to the representation mappings $\hat{g}_{\mathcal{M}}$, $\hat{g}_{\mathcal{N}}$ and $g^{\star}$. $M_i$ denotes modality $i$.}
\label{fig1}
\end{figure}

Under this composite framework, we provide  the first theoretical analysis to shed light on what makes multi-modal outperform uni-modal from the generalization perspective. We identify the relationship between the population risk and the distance between a learned  latent representation $\hat{g}$ and the $g^{\star}$, under the metric we will define later. Informally, closer to the true representation leads to less population loss, which indicates that a better latent representation guarantees the end-to-end multi-modal learning performance. Instead of simply considering the comparison of \textit{multi vs uni} modalities, we consider a general case, \textit{$\mathcal{M}$ vs $\mathcal{N}$} modalities, which are distinct subsets of all modalities. We focus on the condition that the latter is a subset of the former. Our second result is a bound for  the closeness between $\hat{g}$ and the $g^{\star}$, from which we provably show that the latent representation $\hat{g}_{\mathcal{M}}$ learning from the $\mathcal{M}$ modalities is closer to the true  $g^{\star}$ than $\hat{g}_{\mathcal{N}}$ learning from $\mathcal{N}$ modalities.  As shown in Figure~\ref{fig1}, $\hat{g}_{\mathcal{M}}$ has a more sufficient latent space exploration than $\hat{g}_{\mathcal{N}}$. Moreover, in a specific linear regression model, we directly verify that using multiple modalities rather than its subset  learns a better latent representation.

%In section 5, we use experiments on natural dataset and simulated data to verify our theory.

The main contributions of this paper are summarized as follows:
\begin{itemize}
    \item We formalize the multi-modal learning problem into a theoretical framework. Firstly, we show that the performance of multi-modal learning in terms of population risk can be bounded by the \textit{latent representation quality}, a novel metric we propose to measure the distance from a learned latent representation to the true representation, which reveals that the ability of learning the whole task coincides with the ability of learning the latent representation when we have sufficient training samples. %\textcolor{red}{include detailed statements}
    \item We derive an upper bound for the latent representation quality of training over a subset of modalities. This directly implies a principle to guide us in modality selection, i.e., when the number of sample size is large and multiple modalities can efficiently optimize the empirical risk, using multi-modal to build a recognition or detection system can have a better performance. %\longbo{the last sentence is not clear}
    % when that single modality is contained in the  former.%\textcolor{red}{formalize}
    \item Restricted to linear latent and task mapping, we provide rigorous theoretical analysis that latent representation quality degrades when the subset of multiple modalities is applied. Experiments are also carried out to empirically validate the theoretical observation that $\hat{g}_{\mathcal{N}}$ is inferior to $\hat{g}_{\mathcal{M}}$.
\end{itemize}

The rest of the paper is organized as follows. In the next section,
we review the related literature. The formulation of multi-modal learning problem is described in Section~\ref{sec3}. Main results are presented in Section~\ref{sec4}. In Section~\ref{exp}, we show 
simulation results to support our theoretical claims.  Additional discussions about the inefficiency of multi-modal learning are presented in Section~\ref{sec6}. Finally, conclusions are drawn in Section~\ref{sec7}.

\section{Related Work}\label{sec2}

\paragraph{Multi-modal Learning Applications}
Deep learning makes fusing different signals  easier, which enables us to develop many multi-modal frameworks. For example, \cite{seichter2020efficient, jiang2018rednet, hu2019acnet, park2017rdfnet, hazirbas2016fusenet} combine RGB and depth images to improve semantic segmentation; \cite{chen2020vggsound, gemmeke2017audio} fuse audio with video to do scene understanding; researchers also explore  audio-visual source separation and localization \cite{zhao2018sound, ephrat2018looking}.
% 
% Notice that it has been discovered that the use of multi-modal data in practice will degrade the performance of the model in some cases\cite{wang2020makes,han2021trusted,subedar2019uncertainty,gat2020removing}. These works identify the causes of performance drops of multi-modal as interactions between modalities in the training stage and try to improve the performance by proposing new training/optimization strategies. Therefore, to theoretically understanding them, we need to analyze the training process from the optimization perspective, which is orthogonal to our analysis. Understanding why multi-modal fails in practice is a very interesting direction, and worth further investigation in the future research.

\paragraph{Theory of Multi-modal Learning} On the theory side, in semi-supervised setting, \cite{sun2020tcgm} proposed a novel method, Total Correlation Gain Maximization (TCGM), and theoretically
proved that  TCGM can find the groundtruth Bayesian classifier given each modality. Moreover, CPM-Nets \cite{zhang2019cpm} showed multi-view representation can recover the same performance as only using the single-view observation by constructing the versatility. However, they made strict assumptions on the relationship across different modalities, while our analysis does not require such additional assumptions. Besides, previous multi-view analysis~\cite{sridharan2008information, amini2009learning,xu2013survey,federici2020learning} typically assumes that each view alone is sufficient to
predict the target accurately,  %the single view has sufficient information for the learning task,e.g. classification
 which may not hold in our multi-modal setting. For instance, 
it is difficult to build a classifier just using a weak modality with limited labeled data, e.g., depth modality in RGB-D images for object detection task~\cite{gupta2016cross}. %Recently, \cite{du2021improving} analyzed the potential negative aspect of multi-modal learning, and provably showed that multi-modal methods will inevitably fail to learn the features of one of the modalities under the imbalanced multi-modal data hypothesis.
\paragraph{Transfer Learning}
A line of work closely related to our composite learning framework is transfer learning via representation learning, which firstly learns a shared representation on various tasks and then transfers the learned representation to a new task. \cite{tripuraneni2020provable,cavallanti2010linear,pontil2013excess,du2020few} have provided the sample complexity bounds in the special case of the linear feature and linear task mapping.  \cite{tripuraneni2020theory} introduces a new notion of task diversity and provides a generalization bound with general tasks, features, and losses. Unfortunately, the function class which contains feature mappings is the same across all different tasks while our focus is that the function classes generated by different subsets modalities  are usually inconsistent.
\paragraph{Notation:}
%\subsection{Single and Multi Modalities} 
Throughout the paper, we use 
$\|\cdot\|$ to denote the $\ell_2$ norm. We also denote the set of positive integer numbers less or equal than $n$ by $[n]$, i.e. $[n]\triangleq\left\{1,2,\cdots,n\right\}$.

\section{The Multi-modal Learning Formulation}\label{sec3}
In this section, we present the Multi-modal Learning problem formulation. 
% 
%\subsection{Multi-Modality Learning with a Latent Space}
Specifically, we assume that a given data $\mathbf{x}:=\left(x^{(1)},\cdots, x^{(K)}\right)$ consists of $K$ modalities, where $x^{(k)}\in \mathcal{X}^{(k)}$ the domain set of the $k$-th modality. Denote $\mathcal{X}=\mathcal{X}^{(1)}\times\cdots\times\mathcal{X}^{(K)}$. 
We use $\mathcal{Y}$ to denote the target domain and use $\mathcal{Z}$ to denote a latent space. Then, we denote $g^{\star}:\mathcal{X}\mapsto\mathcal{Z}$ the true mapping from the input space  
(using all of $K$ modalities) to the
latent space, and  $h^{\star}:\mathcal{Z}\mapsto\mathcal{Y}$ is the true task mapping. For instance,  in aggregation-based multi-modal fusion,  $g^{\star}$ is an aggregation function compounding on $K$ seperate sub-networks and $h^{\star}$ is a multi-layer neural network~\cite{wang2020deep}. %\longbo{give examples of g and h}

In the learning task, a data pair $(\mathbf{x},y)\in\mathcal{X}\times \mathcal{Y}$ is generated from an unknown distribution $\mathcal{D}$, such that
\begin{align}
    \mathbb{P}_{\mathcal{D}}(\mathbf{x},y)\triangleq \mathbb{P}_{y \mid \mathbf{x}}\left(y \mid h^{\star} \circ g^{\star}(\mathbf{x})\right)\mathbb{P}_{\mathbf{x}}(\mathbf{x})\label{dis}
\end{align}
Here $h^{\star} \circ g^{\star}(\mathbf{x}) = h^{\star} (g^{\star}(\mathbf{x}))$ represents the composite function of $h^{\star}$ and $g^{\star}$. 
%$g^{\star}$ is the mapping from the input space% $\mathcal{X}\triangleq \mathcal{X}^{(1)}\times\cdots\times\mathcal{X}^{(K)}$
%(using all of $K$ modalities) to a %low-dimensional 
%latent space $\mathcal{Z}$. And $h^{\star}:\mathcal{Z}\mapsto\mathcal{Y}$ is the task mapping.

 In real-world settings, we often face incomplete multi-modal data, i.e., some modalities are not observed. 
 To take into account this situation, 
we  let $\mathcal{M}$ be a subset of $[K]$, and without loss of generality,  focus on the learning problem only using the  modalities in  $\mathcal{M}$. Specifically, define $\mathcal{X}^{\prime}:=\left(\mathcal{X}^{(1)} \cup\{\perp\}\right) \times \ldots \times\left(\mathcal{X}^{(K)} \cup\{\perp\}\right)$ as the extension  of $\mathcal{X}$, where  $\mathbf{x}^{\prime}\in \mathcal{X}^{\prime}$,  $\mathbf{x}^{\prime}_{k}=\perp$ means that the $k$-th modality is not used (collected).  Then we define a mapping $p_{\mathcal{M}}$ from $\mathcal{X}$ to $\mathcal{X}^{\prime}$ induced by $\mathcal{M}$:
 $$
 p_{\mathcal{M}}(\mathbf{x})^{(k)}=
 \begin{cases}
  \mathbf{x}^{(k)} & \text{if } k\in \mathcal{M}\\
\perp& \text{else}
 \end{cases}
 $$
 Also define $p^{\prime}_{\mathcal{M}}: \mathcal{X}^{\prime}\mapsto\mathcal{X}^{\prime}$ as the extension of $p_{\mathcal{M}}$. Let $\mathcal{G}^{\prime}$ denote a function class, which contains the mapping from $\mathcal{X}^{\prime}$ to the latent space $\mathcal{Z}$, and define a function class $\mathcal{G}_{\mathcal{M}}$   as follows:
\begin{align}
    \mathcal{G}_{\mathcal{M}}\triangleq \{g_{\mathcal{M}}:\mathcal{X}\mapsto\mathcal{Z}\mid g_{\mathcal{M}}(\mathbf{x}):=g^{\prime}(p_{\mathcal{M}}(\mathbf{x})), g^{\prime}\in\mathcal{G}^{\prime}\}
\end{align}

Given a data set $\mathcal{S}=\left(\left(\mathbf{x}_{i}, y_{i}\right)\right)_{i=1}^{m}$, where $\left(\mathbf{x}_{i}, y_{i}\right)$ is drawn i.i.d. from $\mathcal{D}$, the learning objective is, following the  Empirical Risk Minimization (ERM) principle~\cite{mohri2018foundations}, to find $h\in\mathcal{H}$ and $g_{\mathcal{M}}\in\mathcal{G}_{\mathcal{M}}$ to jointly minimize the empirical risk, i.e., 
\begin{eqnarray}
    \min&&\hat{r}\left(h\circ g_{\mathcal{M}}\right)\triangleq \frac{1}{m}\sum_{i=1}^{m}\ell\left(h\circ g_{\mathcal{M}}(\mathbf{x}_{i}),y_i\right)\\
    \text{s.t.} && h\in\mathcal{H}, g_{\mathcal{M}}\in\mathcal{G}_{\mathcal{M}}. 
\end{eqnarray}
where $\ell(\cdot,\cdot)$ is the loss function. Given $\hat{r}\left(h\circ g_{\mathcal{M}}\right)$, we similarly define its corresponding population risk as
\begin{eqnarray}
r\left(h\circ g_{\mathcal{M}}\right)=\mathbb{E}_{(\mathbf{x}_i,y_i)\sim\mathcal{D}} \left[\hat{r}\left(h\circ g_{\mathcal{M}}\right)\right]    
\end{eqnarray}
Similar to \cite{amini2009learning,tripuraneni2020theory}, we  use the population risk to measure the performance of learning.
%\longbo{explain why use this measure. give refs}

\paragraph{Example.}As a concrete example of our model, consider the video classification problem under the late-fusion model in \cite{wang2020makes}. In this case, each modality $k$, e.g. RGB frames, audio or optical flows, is encoded by a deep network $\varphi_{k}$, and their features are fused and passed to a classifier $\mathcal{C}$. If we train on the first $M$ modalities, we can let $\mathcal{M}=[M]$. Then $g_{\mathcal{M}}$ has the form: $\varphi_{1} \oplus \varphi_{2} \oplus \cdots \oplus \varphi_{M}$, where $\oplus$ denotes a fusion operation, e.g. self-attention ($\mathcal{Z}$ is the output of $g_{\mathcal{M}}$), and $h$ is the  classifier $\mathcal{C}$. More examples are provided in Appendix~\ref{composite framework}.

\paragraph{Why Composite Framework ?}Note that the composite multi-modal framework is often observed in applications. In fact, in recent years,  a large number of papers, e.g.,  \cite{baltruvsaitis2018multimodal,fayek2020large, feichtenhofer2016convolutional, wang2020deep,wang2020makes,kalfaoglu2020late}, appear to have utilized this framework in one way or another, even though the contributors did not clearly summarize the relationship between their methods and this common underlying structure. However, despite the popularity of the framework, existing works lack a very formal definition in theory. 

\paragraph{What is Special about Multi-modal Learning?}
For the multi-modal representation $\mathbf{x}:=\left(x^{(1)}, \cdots, x^{(K)}\right)$ consists of $K$ modalities, we allow the dimension of the domain set of each modality $\mathcal{X}^{(k)}$ to be different, which well models the source of heterogeneity of each modality. The relationships across different modalities are usually of varying levels due to the heterogeneous sources. Therefore, compared to previous works~\cite{xu2013survey,zhang2019cpm}, we make no assumptions on the relationship across every single modality in our analysis, which makes it general to allow different correlations.  Moreover, the main assumption behind previous analysis~\cite{sridharan2008information,sun2020tcgm,tsai2020self} is that each view/modality contains sufficient information for target tasks, which 
    does not shed light on our analysis. It may not hold in multi-modal applications~\cite{yang2015auxiliary}, e.g., in object detecting task, it is known that depth images are with lower accuracy than RGB images~\cite{gupta2016cross}.

%\longbo{give a concrete example to help readers understand the model. also provide refs}

%\longbo{give a concrete example to help readers understand the model. also provide refs}

\section{Main Results}\label{sec4}
In this section, we provide main theoretical results to rigorously establish various aspects of the folklore
claim that multi-modal is better than single. We first detail several assumptions  throughout this section.
\begin{assumption}\label{assump1}
The loss function $\ell(\cdot,\cdot)$ is $L$-smooth with respect to the first coordinate, and is bounded by a constant $C.$
 \end{assumption}
 \begin{assumption}\label{assump3}
 The true latent representation $g^{\star}$ is contained in $\mathcal{G}$, and the task mapping $h^{\star}$ is contained in $\mathcal{H}$.
 \end{assumption}
Assumption \ref{assump1} is a classical regularity condition for loss function in theoretical analysis~\cite{mohri2018foundations,tripuraneni2020theory,tripuraneni2020provable}. 
Assumption~\ref{assump3} is also be known as realizability condition in representation learning~\cite{tripuraneni2020theory,du2020few,tripuraneni2020provable}, which ensures that the function class that we optimize over  contains the true latent representation and the task mapping. 
  \begin{assumption}\label{assump2}
 For any $g^{\prime}\in\mathcal{G}^{\prime}$ and $\mathcal{M}\subset [K]$,  $g^{\prime}\circ p_{\mathcal{M}}^{\prime}\in \mathcal{G}^{\prime}$. 
 \end{assumption}
  To understand Assumption \ref{assump2}, note that for any $\mathcal{N}\subset\mathcal{M}\subset [K]$, by definition, for any $g_{\mathcal{N}}\in\mathcal{G}_{\mathcal{N}}$, there exists  $g^{\prime}\in \mathcal{G}^{\prime}$, s.t. $$g_{\mathcal{N}}(\mathbf{x})=g^{\prime}(p_{\mathcal{N}}(\mathbf{x}))=g^{\prime}(p_{\mathcal{N}}^{\prime}(p_{\mathcal{M}}(\mathbf{x})))$$
 Therefore, Assumption~\ref{assump2} directly implies $g_{\mathcal{N}}\in \mathcal{G}_{\mathcal{M}}$. Moreover, we have $\mathcal{G}_{\mathcal{N}}\subset \mathcal{G}_{\mathcal{M}}\subset \mathcal{G}$, which means that the inclusion relationship of modality subsets remains unchanged on the latent function class induced by them. As an example, if $\mathcal{G}^{\prime}$ is linear, represented as matrix $\mathbf{G}\in\mathbb{R}^{Q\times K}$.  Also $p^{\prime}_{\mathcal{M}}$ can be represented as a diagonal matrix $\mathbf{P}\in \mathbb{R}^{K\times K}$ with the $i$-th diagonal entry being $1$ for $i\in\mathcal{M}$ and $0$ otherwise. In this case, Assumption~\ref{assump2} holds, i.e. $\mathbf{G}\times\mathbf{P}\in \mathcal{G}^{\prime}$. Moreover, $\mathbf{G}\times\mathbf{P}$ is a matrix with $i$-th column all be zero for $i\notin\mathcal{M}$, which is commonly used in the underfitting analysis in linear regression~\cite{seber2012linear}.
 % 
% \longbo{give the linear example to explain}
 %
 
% \longbo{explain the assumptions and give refs}
 
 \subsection{Connection to Latent Representation Quality}
Latent space is employed to better exploit the correlation among different modalities. Therefore, we will naturally conjecture that the performance of training with different modalities is related to its ability to learn latent space representation. In this section, we will formally characterize this relationship. 

In order to measure the goodness of a learned latent representation $g$,  we   introduce the following definition of \textit{latent representation quality}. 
\begin{definition}{1}
Given a data distribution with the form in $(\ref{dis})$,  for any learned latent representation mapping $g\in\mathcal{G}$, the \textbf{latent representation quality} is defined as
\begin{align}
   \eta(g) = \inf_{h\in \mathcal{H}}\left[ r\left(h\circ g\right)-r(h^{*}\circ g^{*})\right]
\end{align}
\end{definition}
Here $\inf_{h\in \mathcal{H}} r\left(h\circ g\right)$ is the best achievable population risk with the fixed latent representation $g$. Thus, to a certain extent, $\eta(g)$ measures the loss incurred  by the distance between $g$ and $g^{\star}$. 

%\subsection{Model Complexity}

Next, we recap the Rademacher complexity measure for  model complexity. It will be used in quantifying the population risk performance based on different modalities. 
Specifically, let $\mathcal{F}$ be a class of vector-valued function $\mathbb{R}^{d} \mapsto \mathbb{R}^{n}$. Let $Z_{1}, \ldots, Z_{m}$ be i.i.d. random variables on $\mathbb{R}^{d}$ following some distribution $P .$  Denote the sample $S=\left(Z_{1}, \ldots, Z_{m}\right)$. The empirical Rademacher complexity of $\mathcal{F}$ with respect to the sample $S$ is given by~\cite{bartlett2002rademacher} % 
$$
\widehat{\mathfrak{R}}_{S}(\mathcal{F}):=\mathbb{E}_{\sigma}\left[\sup _{f \in \mathcal{F}} \frac{1}{m} \sum_{i=1}^{m} \sigma_{i} f\left(Z_{i}\right)\right]
$$
where $\sigma=\left(\sigma_{1}, \ldots, \sigma_{n}\right)^{\top}$ with $\sigma_{i} \stackrel{i i d}{\sim}$ unif $\{-1,1\} .$ The Rademacher complexity of $\mathcal{F}$ is
$$
\mathfrak{R}_{m}(\mathcal{F})=\mathbb{E}_{S}\left[\widehat{\mathfrak{R}}_{S}(\mathcal{F})\right]
$$

Now we present our first main result regarding multi-modal learning. 
\begin{theorem}\label{thm:mn-modality}
 Let $\mathcal{S}=\left(\left(\mathbf{x}_{i}, y_{i}\right)\right)_{i=1}^{m}$ be a dataset of $m$ examples drawn i.i.d. according to $\mathcal{D} .$ Let $\mathcal{M},\mathcal{N}$ be two distinct subsets of $ [K]$. Assuming we have produced the empirical risk minimizers $(\hat{h}_{\mathcal{M}},\hat{g}_{\mathcal{M}})$ and $(\hat{h}_{\mathcal{N}},\hat{g}_{\mathcal{N}})$, training with the $\mathcal{M}$ and $\mathcal{N}$  modalities separately. Then, 
        for all $1>\delta>0,$ with probability at least $1-\frac{\delta}{2}$:
        \begin{align}
&r\left(\hat{h}_{\mathcal{M}}\circ \hat{g}_{\mathcal{M}}\right)- r\left(\hat{h}_{\mathcal{N}}\circ \hat{g}_{\mathcal{N}}\right)\nonumber\\
&\leq \gamma_{\mathcal{S}}(\mathcal{M},\mathcal{N})+8L\mathfrak{R}_{m}( \mathcal{H}\circ\mathcal{G}_{\mathcal{M}})+\frac{4C}{\sqrt{m}}+2C\sqrt{\frac{2 \ln (2 / \delta)}{m}}\label{mg}
\end{align}
          where 
    \begin{align}&\gamma_{\mathcal{S}}(\mathcal{M},\mathcal{N})\triangleq \eta(\hat{g}_{\mathcal{M}})-\eta(\hat{g}_{\mathcal{N}})\label{gamma} \qquad\Box
    \end{align}
    \end{theorem}

\textbf{Remark.} 
A few remarks are in place. First of all, $\gamma_{\mathcal{S}}(\mathcal{M},\mathcal{N})$  defined  in $(\ref{gamma})$ compares the quality between  latent representations learning from $\mathcal{M}$ and $\mathcal{N}$ modalities with respect to the given dataset $\mathcal{S}$. Theorem \ref{thm:mn-modality}  bounds the difference of population risk training with two different subsets of modalities by $\gamma_{\mathcal{S}}(\mathcal{M},\mathcal{N})$, which validates our conjecture that including more modalities is advantageous in learning.
  Second, for the commonly used function classes in the field of machine learning,  Radamacher complexity for a sample of size $m$, $\mathfrak{R}_{m}(\mathcal{F})$ is usually bounded by  $\sqrt{C(\mathcal{F}) / m}$, where $C(\mathcal{F})$ represents the intrinsic property of function class $\mathcal{F}$. 
  Third,   $(\ref{mg})$ can be written as $\gamma_{\mathcal{S}}(\mathcal{M},\mathcal{N})+\mathcal{O}(\sqrt{\frac{1}{m}})$ in order terms. This shows that as the number of sample size grows, the performance of using different modalities mainly depends on its latent representation quality. %\longbo{these insights are interesting. should highlight in the contributions}
%\end{remark}

\subsection{Upper Bound for Latent Space Exploration}
Having establish the connection between the population risk difference with  latent representation quality,  our next goal is to estimate how close the learned latent representation $\hat{g}_{\mathcal{M}}$ is to  the true latent representation  $g^{\star}$. The following theorem shows how the latent representation quality can be controlled in the training process.
\begin{theorem}\label{thm-latent}
 Let $\mathcal{S}=\{\left(\mathbf{x}_{i}, y_{i}\right)\}_{i=1}^{m}$ be a dataset of $m$ examples drawn i.i.d. according to $\mathcal{D} .$ Let $\mathcal{M}$ be a subset of $ [K]$. Assuming we have produced the empirical risk minimizers $(\hat{h}_{\mathcal{M}},\hat{g}_{\mathcal{M}})$ training with the $\mathcal{M}$ modalities. Then,  
        for all $1>\delta>0,$ with probability at least $1-\delta$:
 \begin{align}
     \eta(\hat{g}_{\mathcal{M}})\leq
     4L\mathfrak{R}_{m}( \mathcal{H}\circ\mathcal{G}_{\mathcal{M}})+4L\mathfrak{R}_{m}( \mathcal{H}\circ\mathcal{G})%+\frac{4C}{\sqrt{m}}
     +6C\sqrt{\frac{2 \ln (2 / \delta)}{m}}
     +\hat{L}(\hat{h}_{\mathcal{M}}\circ\hat{g}_{\mathcal{M}},\mathcal{S})
\end{align}
where $\hat{L}(\hat{h}_{\mathcal{M}}\circ\hat{g}_{\mathcal{M}},\mathcal{S})\triangleq \hat{r}\left(\hat{h}_{\mathcal{M}}\circ \hat{g}_{\mathcal{M}}\right)-\hat{r}\left(h^{\star}\circ g^{\star}\right)$ is the centered empirical loss. $\Box$
\end{theorem}

\textbf{Remark.} 
%From Theorem \ref{thm-latent}, we see that for 
%
Consider sets $\mathcal{N}\subset\mathcal{M}\subset [K]$. 
Under Assumption~\ref{assump2}, $\mathcal{G}_{\mathcal{N}}\subset \mathcal{G}_{\mathcal{M}}\subset \mathcal{G}$, optimizing over a larger function class results in a smaller empirical risk. Therefore
    \begin{align}
    \hat{L}(\hat{h}_{\mathcal{M}}\circ\hat{g}_{\mathcal{M}},\mathcal{S})\leq \hat{L}(\hat{h}_{\mathcal{N}}\circ\hat{g}_{\mathcal{N}},\mathcal{S})
        %\min_{h\in\mathcal{H},g\in\mathcal{G}_{\mathcal{M}}}\hat{r}\left(h\circ g\right)\leq  \min_{h\in\mathcal{H},g\in\mathcal{G}_{\mathcal{N}}}\hat{r}\left(h\circ g\right)
    \end{align}
 Similar to the analysis in Theorem~\ref{thm:mn-modality}, the first term  on the  Right-hand Side (RHS),  $\mathfrak{R}_{m}( \mathcal{H}\circ\mathcal{G}_{\mathcal{M}})\sim \sqrt{C(\mathcal{H}\circ\mathcal{G}_{\mathcal{M}})/m}$ and $\mathfrak{R}_{m}( \mathcal{H}\circ\mathcal{G}_{\mathcal{N}})\sim \sqrt{C(\mathcal{H}\circ\mathcal{G}_{\mathcal{N}})/m}$. Following the basic structural property of Radamacher complexity~\cite{bartlett2002rademacher}, we have $C(\mathcal{H}\circ\mathcal{G}_{\mathcal{N}})\leq C(\mathcal{H}\circ\mathcal{G}_{\mathcal{M}})$.
%\end{itemize}
Therefore, Theorem \ref{thm-latent} offers the following principle for choosing modalities to improve the latent representation quality. 

%sheds light on how more modalities can improve the \textit{latent representation quality}: 
\textbf{Principle:} \textit{choose to learn with more modalities if:}
$$
\hat{L}(\hat{h}_{\mathcal{N}}\circ\hat{g}_{\mathcal{N}},\mathcal{S}) - \hat{L}(\hat{h}_{\mathcal{M}}\circ\hat{g}_{\mathcal{M}},\mathcal{S})\geq \sqrt{\frac{C(\mathcal{H}\circ\mathcal{G}_{\mathcal{M}})}{m}}- \sqrt{\frac{C(\mathcal{H}\circ\mathcal{G}_{\mathcal{N}})}{m}} $$ 
%\longbo{highlight this in conrributions}
What this principle implies are twofold. (i) When the  number of sample size $m$ is large, the impact of intrinsic complexity of function classes will be reduced. (ii) Using more modalities can efficiently optimize the empirical risk, hence improve 
the latent representation quality.%\longbo{rewrite this sentence} 

Through the trade-off illustrated in the above principle,  we provide theoretical evidence that when $\mathcal{N}\subset\mathcal{M}$ and training samples are sufficient, $\eta(\hat{g}_{\mathcal{M}})$ may be less than $\eta(\hat{g}_{\mathcal{N}})$, i.e.$ \gamma_{\mathcal{S}}(\mathcal{M},\mathcal{N})\leq 0$. Moreover, combining with the  conclusion from Theorem~\ref{thm:mn-modality}, if the sample size $m$ is large enough, $ \gamma_{\mathcal{S}}(\mathcal{M},\mathcal{N})\leq 0$ guarantees $r\left(\hat{h}_{\mathcal{M}}\circ \hat{g}_{\mathcal{M}}\right)\leq r\left(\hat{h}_{\mathcal{N}}\circ \hat{g}_{\mathcal{N}}\right)$, which indicates learning with the $\mathcal{M}$ modalities outperforms only using its subset  $\mathcal{N}$ modalities.

\paragraph{Role of the intrinsic property $\mathcal{C}(\cdot)$}
Hypothesis function classes are typically overparametrized in deep learning, and will be extremely large for some typical measures, e.g., VC dimension, absolute dimension. Some recent efforts aim to offer an explanation about why neural networks generalize better with over-parametrization~\cite{neyshabur2018towards, neyshabur2015norm, bartlett2017spectrally}. \cite{neyshabur2018towards} suggest a novel complexity measure based on unit-wise capacities, which implies if both in overparametrized settings, the complexity will not change much or even decrease when we have more modalities (using more parameters). Thus, the inequality in the principle is trivially satisfied, and we will choose to learn with more modalities.

%\paragraph{Take Away Message:} The key takeaway message in this section is that the success of multi-modal learning relies essentially on the better quality of  latent space representation. %Specifically, we prove that the performance of multi-modal learning in terms of population risk can be bounded by the latent representation quality, and provide an upper bound for the latent representation quality of training over a subset of modalities. 

%Utilizing the result of Theorem 1, the non-positivity of $ \gamma_{\mathcal{S}}(\mathcal{M},\mathcal{N})$ indicates learning with the $\mathcal{M}$ modalities outperforms only using its subset  $\mathcal{N}$ modalities.\longbo{this last statement is weak}

%Utilizing the result of Theorem 1, the non-positivity of $ \gamma_{\mathcal{S}}(\mathcal{M},\mathcal{N})$ indicates learning with the $\mathcal{M}$ modalities outperforms only using its subset  $\mathcal{N}$ modalities.\longbo{this last statement is weak}

\subsection{Non-Positivity Guarantee} 
In this section, we focus on a composite linear data generating model to theoretically verify that the $\gamma_{\mathcal{S}}(\mathcal{M},\mathcal{N})$ is indeed non-positive in this special case.\footnote{Proving that $\gamma_{\mathcal{S}}(\mathcal{M},\mathcal{N})\leq0$ holds in general is open and will be an interesting future work.}%which can provide very strong theoretical guarantee for multi-modal learning.\longbo{polish}} 
%\longbo{ a bit confusing. is it non-positive in general or only in this special case?} 
% 
Specifically, we consider the case where the mapping to the latent space and the task mapping are both linear. Formally, let the function class $\mathcal{G}$ and $\mathcal{H}$ be:
\begin{equation}
\begin{array}{l}
\mathcal{G}=\left\{g \mid g(\mathbf{x})=\mathbf{A}^{\top} \mathbf{x}, \mathbf{A} \in \mathbb{R}^{d \times n}, \mathbf{A}  \right\}\\
\mathcal{H}=\left\{h \mid h(\mathbf{z})=\boldsymbol{\beta}^{\top} \mathbf{z}, \boldsymbol{\beta} \in \mathbb{R}^{n},\|\boldsymbol{\beta}\| \leq C_b\right\} 
\end{array}
\end{equation}
where $\mathbf{x}=\left(\mathbf{x}^{(1)},\cdots, \mathbf{x}^{(K)}\right)$ is a $d$-dimensional vector, $\mathbf{x}^{(k)}\in \mathbb{R}^{d_{k}}$ denotes the feature vector for the $k$-th modality and $\sum_{k=1}^{K}d_k=d$. Here, the distribution $\mathbb{P}_{\mathbf{x}}(\cdot)$ satisfies that its  covariance matrix is positive definite. The data is generated by:
\begin{align}
    y=\left(\boldsymbol{\beta}^{\star}\right)^{\top} {\mathbf{A}^{\star}}^{\top}\mathbf{x}+\epsilon\label{ldg}
\end{align}
where r.v. $\epsilon$ is independent of $\mathbf{x}$ and has zero-mean and bounded second moment. 
Note that in practical multi-modal learning, usually only one layer is  linear and the other is a neural network. 
For instance, 
\cite{zhou2020multi} employs the linear matrix to project the feature matrix from different modalities into a common latent space for early dementia diagnosis, i.e., $\mathcal{G}$ is linear. Another example is in pedestrian detection \cite{xu2017learning}, where a linear task mapping is adopted, i.e., $\mathcal{H}$ is linear.
% \longbo{give examples to what this model corresponds}
Thus, our composite linear model can be viewed as an approximation to such popular models, and our results can offer insights into the performance of these models.

We consider a special case that $\mathcal{M}=[K]$ and $\mathcal{N} = [K-1]$. Thus  $\mathcal{G}_{\mathcal{M}}=\mathcal{G}$ and we have the following result. 
\begin{equation}
\begin{array}{l}
\mathcal{G}_{\mathcal{N}}=\left\{g \mid g(\mathbf{x})=\left[\begin{array}{cc}\mathbf{A}_{1:\sum_{k=1}^{K-1}d_k}\\ \mathbf{0}\end{array}\right]^{\top} \mathbf{x}, \mathbf{A} \in \mathbb{R}^{d \times n}\text { with orthonormal columns }\right\}
\end{array}
\end{equation}
\begin{proposition}\label{prop-non-positive}
Consider the dataset $\mathcal{S}=\{\left(\mathbf{x}_{i}, y_{i}\right)\}_{i=1}^{m}$ generating from the linear model defined in $(\ref{ldg})$ with $\ell_2$ loss. Let $\mathcal{M}=[K]$ and $\mathcal{N} = [K-1]$.    Let $\hat{\mathbf{A}}_{\mathcal{M}}$, $\hat{\mathbf{A}}_{\mathcal{N}}$ denote the projection matrix estimated by $\mathcal{M}$, $\mathcal{N}$ modalities. Assume that $\hat{\mathbf{A}}_{\mathcal{M}}$, $\mathbf{A}^{\star}$ has orthonormal columns.  If $n=d$, for sufficiently large constant $C_b$, we have:
\begin{align}
\gamma_{\mathcal{S}}(\mathcal{M},\mathcal{N})\leq 0
\end{align}
\end{proposition}

In this special case, Proposition \ref{prop-non-positive} directly guarantees that  training with incomplete modalities weakens the ability to learn a  optimal latent representation. As a result, it also  degrades the learning performance.

\section{Experiment}\label{exp}
We conduct experiments to validate our theoretical results. The source of the data we consider is two-fold, multi-modal real-world dataset and well-designed generated dataset. 
% All experiments are based on open source machine learning library PyTorch \cite{paszke2019pytorch}.

%on multi-modal natural dataset and simulation  

\subsection{Real-world dataset}
%In this subsection, we show results on Natural Dataset to verify our theory.
\paragraph{Dataset.} The natural dataset we use is the  Interactive Emotional Dyadic Motion Capture (IEMOCAP) database, which is an acted multi-modal and multi-speaker database~\cite{busso2008iemocap}. It contains three modalities, Text, Video and Audio. We follow the data preprocessing method of \cite{poria2017context} and obtain $100$ dimensions data for audio,   $100$ dimensions for text, and   $500$ dimensions for video. There are six labels here, namely, happy, sad, neutral, angry, excited and frustrated. We use $13200$ data for training and $3410$ for testing. 

\paragraph{Training Setting.} For all experiments on IEMOCAP, we use one linear neural network layer to extract the latent feature, and we set the hidden dimension to be $128$. In multi-modal network, different modalities do not share encoders and we concatenate the features first, and then map the feature to the task space. We use Adam \cite{kingma2014adam} as the optimizer and set the learning rate to be $0.01$, with other hyper-parameters default. The batch size is $2048$ for the data.  For this classification task, the top-1 accuracy is used for performance measurement. We use naively multi-modal late-fusion training as our framework~\cite{wang2020makes, du2021improving}. In Appendix~\ref{appendix:training_setting}, we provide more discussions on stable multi-modal training.

% \du{When performing multi-modal joint training, models are easy to learn insufficient representations of each modality~\cite{du2021improving}. To avoid this problem and ensure stable training, we first perform uni-modal training to get trained encoders of each modality and then train a new multi-modal classifier over them in Table~\ref{tab:data} and \ref{tab:ienum}.}

\paragraph{Connection to the Latent Representation Quality.} The classification accuracy on IEMOCAP, using different combinations of modalities are summarized in Table~\ref{tab:data}. All learning strategies using multiple modalities outperform
the single modal baseline. To validate Theorem~\ref{thm:mn-modality}, we calculate the test accuracy difference  between different subsets of modalities using the result in Table~\ref{tab:data} and show them in the third column of Table~\ref{tab:com}. 

Moreover, we empirically evaluate the $\eta(\hat{g}_{\mathcal{M}})$ in the following way: freeze the encoder $\hat{g}_{M}$ obtained through  pretraining and then finetune to obtain a better classifier $h$. Having  $\eta(\hat{g}_{\mathcal{M}})$ and $\eta(\hat{g}_{\mathcal{N}})$, the values of $\gamma_{\mathcal{S}}(\mathcal{M},\mathcal{N})$ between different subsets of modalities are also presented in Table~\ref{tab:com}. Previous discussions on Theorem~\ref{thm:mn-modality} implies that the population risk difference between $\mathcal{M}$ and $\mathcal{N}$ modalities has the same sign as $\gamma_{\mathcal{S}}(\mathcal{M},\mathcal{N})$ when the sample size is large enough, and negativity implies performance gains. Since we use accuracy as the measure, on the contrary, positivity indicates a better performance in our settings. As shown in Table~\ref{tab:com}, when more modalities are added for learning, the test accuracy difference and $\gamma_{\mathcal{S}}(\mathcal{M},\mathcal{N})$ are both positive, which confirms the important role of latent representation quality characterized in Theorem~\ref{thm:mn-modality}.

\begin{table}[ht]
    \centering
    \begin{tabular}{cc}
       \toprule
 Modalities & Test Acc  \\
 \hline
 Text(T) & 49.93$\pm$0.57 \\
 Text + Video(TV)& 51.08$\pm$0.66 \\
 Text + Audio(TA) & 53.03$\pm$0.21 \\
 Text + Video + Audio(TVA)& \textbf{53.89 $\pm$ 0.47} \\
 \bottomrule
    \end{tabular}
    \caption{Test classification accuracy on IEMOCAP, using different combinations of modalities, only Text, Text + Video, Text + Audio and  Text + Video + Audio.}\label{tab:data}
\end{table}

\begin{table}[h]
   \centering
\begin{tabular}{cccccc}
 \toprule
  \cr \multicolumn{1}{c}{Modalities}   & \multicolumn{5}{c}{ Test Acc (Ratio of Sample Size)} \cr
 \cmidrule(lr){2-6}
 & $10^{-4}$ & $10^{-3}$&$10^{-2}$&$10^{-1}$& 1\cr
\midrule
 T   & 23.66$\pm$1.28&29.08$\pm$3.34& 45.63$\pm$0.29&48.30$\pm$1.31& 49.93$\pm$0.57\cr
 TA &  \textbf{25.06$\pm$1.05}&34.28$\pm$4.54 &\textbf{47.28$\pm$1.24}&50.46$\pm$0.61& 51.08$\pm$0.66\cr
 TV &  24.71$\pm$0.87&\textbf{38.37$\pm$3.12} &46.54$\pm$1.62& 49.50$\pm$1.04&53.03$\pm$0.21\cr
 TVA & 24.71$\pm$0.76&32.24$\pm$1.17& 46.39$\pm$3.82&\textbf{50.75$\pm$1.45}&\textbf{53.89$\pm$0.47}\cr
 \bottomrule
\end{tabular}
\caption{Latent representation quality vs. The number of the sample size on IEMOCAP. Noting that in this table, we show the results from naively end-to-end late-fusion training and in Appendix~\ref{appendix:training_setting}, we discuss on more stable multi-modal training methods.}
\label{tab:ienum}
\end{table}

\begin{table}[h]
    \centering
    \begin{tabular}{cccc}
       \toprule
 $\mathcal{M}$ Modalities & $\mathcal{N}$ Modalities  & Test Acc \textbf{Difference}  & $\gamma_{\mathcal{S}}(\mathcal{M},\mathcal{N})$\\
 \hline
 TA &T& 1.15 &1.36\\
 TV &T& 3.10 &3.57\\
 TVA &TV& 0.86& 0.19\\
 TVA& TA& 2.81 &2.4\\
 \bottomrule
    \end{tabular}
    \caption{Comparison of test accuracy and  latent representation quality among different combinations of modalities.}
    \label{tab:com}
\end{table}
\paragraph{Upper Bound for Latent Space Exploration.} Table~\ref{tab:com} also confirms our theoretical analysis in Theorem~\ref{thm-latent}. In all cases,  the $\mathcal{N}$ modalities is a subset of  $\mathcal{M}$ modalities, and correspondingly, a positive $\gamma_{\mathcal{S}}(\mathcal{M},\mathcal{N})$ is observed. This indicates that $\mathcal{M}$ modalities has a more sufficient latent space exploration than  its subset  $\mathcal{N}$ modalities. 

\begin{figure}[h]
  \centering
  \subfigure[Latent representation quality vs.The ratio of the sample size on IEMOCAP]{
    \label{fig:subfig:1} 
    \includegraphics[scale=0.47]{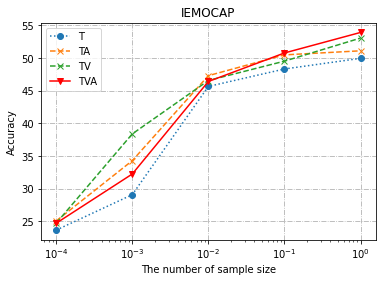}}
  \hspace{0.3in} 
  \subfigure[Latent representation quality vs. The number of modalities on simulated data]{
    \label{fig:subfig:2} 
    \includegraphics[scale=0.47]{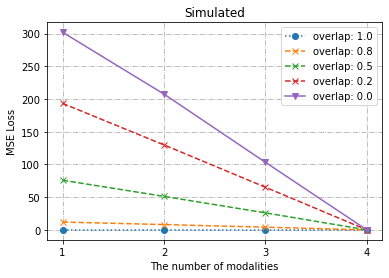}}
  \caption{Evaluation of the latent representation quality on different data sets}
  \label{fig:twopicture} 
\end{figure}

We also attempt to understand the use of sample size for exploring the latent space. Table~\ref{tab:ienum} presents the latent representation  quality $\eta$ obtained by using different numbers of sample size, which is measured by the test accuracy of the pretrain+finetuned modal. Here, the ratio of sample size 
is set to the total number of training samples. The corresponding curve is also ploted in Figure~\ref{fig:subfig:1}. As the number of sample size grows, the increase in performance of $\eta$ is observed, which is in keeping with the $\mathcal{O}(\sqrt{1/m})$ term in our upper bound for $\eta$. The phenomenon that the combination of Text, Video and Audio (TVA) modalities  underperforms the uni-modal when the number of sample size is relatively small, can also be interpreted by the trade-off we discussed in Theorem~\ref{thm-latent}. When there are insufficient
training examples, the intrinsic complexity of the function class induced by multiple modalities dominates, thus weakening its latent representation  quality. 
% to do

%\begin{figure}
%    \centering
%    \includegraphics[width=.5\linewidth]{exp_plot.png}
%    \caption{\textit{Latent representation quality} vs. the number of the sample size on IEMOCAP}
%\end{figure}

%\paragraph{Results.} Only text: 50.03; text + audio: 53.48; text + video: 51.45; text + video + audio: 53.48
%Pretrain + Finetune Classifier: text: 50.03; text + audio: 53.60; text + video: 51.39; text + video + audio: 53.79

\subsection{Synthetic Data}
 In this subsection, we investigate the effect of modality correlation on latent representation quality. Typically, there are three situations for the correlation across each modality~\cite{sun2020tcgm}. %\longbo{ref?}.
 (i) Each modality does not share information at all, that is, each modality only contains modal-specific information. (ii) The other is that, all modalities only maintain the share information without unique information on their own. (iii) The last is a moderate condition, i.e., each modal not only shares information, but also owns modal-specific information. 
 The reason to utilize simulated data in this section is due to the fact that it is hard in practice to have 
natural datasets that possess the required degree of modality correlation. 

%To this end, simulated data is utilized. 

\paragraph{Data Generation.}  %We generate data according to all the three scenarios. %To make each modality only contains the specific information, we generate each modality data by sampling from a normal distribution with mean $0$ and variance $1$. %It is worth noting that each dimension is unrelated. 
%To make each modality do not have share information any more, we sample modality one first, and clone modality one to others to make each modality is totally the same. For the third situation, 
For the synthetic data,  we have four modalities, denoted by $m_1,m_2,m_3,m_4$, respectively,  and the generation process is summarized as follows:
\begin{itemize}
    \item[Step 1:] Generate $m_i\sim \mathcal{N}(\mathbf{0}, \mathbf{I})$, $i=1,2,3,4$,  where $m_i$ is i.i.d. 100-dimensional random vector;
    \item[Step 2:] Generate $m_i\leftarrow (1-w)\cdot m_i+w\cdot m_1$for $i=2,3,4$ 
    \item[Step 3:] Generate the labels as follows. First, add the four modality vectors and calculate the sum of coordinates. Then, obatin the 1-dimensional label, i.e. $y=(m_1+m_2+m_3+m_4).sum(dim=1)$
\end{itemize}
We can see that information from $m_1$ is shared across different modalities,  and $w$ controls how much is shared, which one can use to measure the degree of overlaps. We tune it in $\{0.0,0.2,0.5,0.8,1.0\}$.  A high weight $w$ (close to $1$) indicates a high degree of overlaps, while $w=0$ means each modality is totally independent  and  is non-overlapping.

\paragraph{Training Setting.} We use the multi-layer perceptron neural network as our model. To be more specific, we first use a linear layer to encode the input to a $10$-dimension latent space and then we map the latent feature to the output space. The first layer's input dimension depends on the number of modality. For example, if we use two modalities, the input dimension is $200$. We use SGD as our optimizer, and  set a learning rate $0.01$,  momentum $0.9$, batch size $10000$, training for $10000$ steps. The Mean Square Error (MSE) loss is considered for evaluation in this regression problem.

\paragraph{$\eta$ vs Modality Correlation}
Our aim is to discover the influence of modality correlation on the  latent representation  quality $\eta$. To this end, Table~\ref{tab:sim} shows the  $\eta$ with the varying number of modalities under different correlation conditions, which is measured by the MSE loss of the pretrain+finetuned modal. The trend that the loss decreases as the number of modalities increases is described in Figure~\ref{fig:subfig:2}, which  also validates our analysis of Theorem~\ref{thm-latent}. Moreover,  Figure~\ref{fig:subfig:2} shows that higher correlation among modalities achieves a 
lower loss for $\eta$, which means a better latent representation. This emphasizes the role of latent space to exploit the intrinsic correlations among different modalities. 
%to do

%\paragraph{Results.}
%All modalities share: 
%1: 0
%2: 0
%3: 0
%4: 0

%Only Specific:
%1: 301
%2: 193
%3: 104
%4: 0

%Overlap and Specific:
%1: 77
%2: 50
%3: 27
%4: 0

%\begin{table*}
%\begin{floatrow}
%\capbtabbox{
% \begin{tabular}{cc}
% \hline
% Modalities & Test Acc \\
% \hline
% TV-T & 1.40 \\
% TA-T & 3.45 \\
% TVA-TA & 1.48 \\
% TVA-TV & 3.51 \\
% \hline
% \end{tabular}
%}{
% \caption{IEMOCAP.}
% \label{tab:iemocap}
%}
%\capbtabbox{
% \begin{tabular}{cc}
% \hline
% Modalities & Test  \\
% \hline
% TV-T & 1.36\\
% TA-T & 3.57\\
% TVA-TA &0.19\\
% TVA-TV &2.4\\
% 
% \hline
% \end{tabular}
%}{
% \caption{Optimal $f$.}
% \label{tab:simulated}
%}
%\end{floatrow}
%\end{table*}
\begin{table}[h]
    \centering
    \begin{tabular}{c c c c c c}
     \toprule
     \cr \multicolumn{1}{c}{Modalities}   & \multicolumn{5}{c}{ MSE Loss (Degree of Overlap )} \cr
 \cmidrule(lr){2-6}
          &  1&  0.8 &  0.5 & 0.2& 0.0\\
     \midrule
         $m_1$&0 &12.04$\pm$0.39& 75.89$\pm$1.28&193.28$\pm$1.08& 301.92$\pm$7.85\\
         $m_1, m_2$&0 & 8.16$\pm$0.17& 51.25$\pm$1.06&129.81$\pm$4.36& 207.45$\pm$4.68\\
         $m_1,m_2,m_3$ &0 & 4.18$\pm$0.05&26.06$\pm$0.69 & 65.17$\pm$1.52&103.23$\pm$0.61\\
         $m_1,m_2,m_3,m_4$ &0 & 0 & 0&0&0\\
     \bottomrule
    \end{tabular}
    \caption{Latent representation quality among different correlation situations on synthetic data}
    \label{tab:sim}
\end{table}
\section{Discussion}\label{sec6}
It has been discovered that the use of multi-modal data in practice will degrade the performance of the model in some cases\cite{wang2020makes,han2021trusted,subedar2019uncertainty,gat2020removing}. These works identify the causes of performance drops of multi-modal as interactions between modalities in the training stage and try to improve the performance by proposing new optimization strategies. Therefore, to theoretically understand them, we need to analyze the training process from the \textit{optimization} perspective. %which is orthogonal to our generalization analysis.
Our results, on the other hand, mainly focus on the generalization side, which is separated from optimization and assumes that we get the best performance possible in training. Moreover, our theory is general and does not require additional assumptions on the relationship across every single modality, which may be crucial for theoretically analyzing the observations in a multi-modal performance drop.
Understanding why multi-modal fails in practice is a very interesting direction and worth further investigation in future research.
\section{Conclusion}\label{sec7}
In this work, we formulate a multi-modal learning framework that has been extensively studied in the empirical literature towards rigorously understanding why multi-modality outperforms single since the former has access to a better latent space representation. 
The results answer the two questions: when and why multi-modal outperforms uni-modal  %\longbo{specify what questions} 
jointly. To the best of our knowledge, this is the first theoretical treatment to  explain the superiority of multi-modal from the generalization standpoint.
% future work

The key takeaway message is that the success of multi-modal learning relies essentially on the better quality of  latent space representation, which points to an exciting direction that is worth further investigation:
 to find which encoder is the bottleneck and focus on improving it. More importantly, our work provides new insights for multi-modal theoretical research, and we hope this can encourage more related research.

 \section*{Acknowledgments and Disclosure of Funding} 
 The  work of Yu Huang and Longbo Huang is  supported  in  part  by  the  Technology  and Innovation  Major  Project  of  the  Ministry  of  Science  and Technology  of  China  under  Grant  2020AAA0108400  and 2020AAA0108403.

\bibliographystyle{plain}
\bibliography{main}

\appendix
\appendixpage
\section{Proof of Main Results}
\subsection{Proof of Theorem~\ref{thm:mn-modality}}
\begin{proof}
Let $h^{\prime}_{\mathcal{M}}$ denote the minimizer of the population risk over $\mathcal{D}$ with the representation $\hat{g}_{\mathcal{M}}$, then we can decompose the difference between $r\left(\hat{h}_{\mathcal{M}}\circ \hat{g}_{\mathcal{M}}\right)- r\left(\hat{h}_{\mathcal{N}}\circ \hat{g}_{\mathcal{N}}\right)$ into two parts:
\begin{align}
   & r\left(\hat{h}_{\mathcal{M}}\circ \hat{g}_{\mathcal{M}}\right)- r\left(\hat{h}_{\mathcal{N}}\circ \hat{g}_{\mathcal{N}}\right)\\
    &=\underbrace{ r\left(\hat{h}_{\mathcal{M}}\circ \hat{g}_{\mathcal{M}}\right)- r\left(h^{\prime}_{\mathcal{M}}\circ \hat{g}_{\mathcal{M}}\right)}_{J_1}+\underbrace{r\left(h^{\prime}_{\mathcal{M}}\circ \hat{g}_{\mathcal{M}}\right)- r\left(\hat{h}_{\mathcal{N}}\circ \hat{g}_{\mathcal{N}}\right)}_{J_2}
\end{align}
 $J_1$ can further be decomposed into:
 \begin{align}
     J_1=&\underbrace{r\left(\hat{h}_{\mathcal{M}}\circ \hat{g}_{\mathcal{M}}\right)-\hat{r}\left(\hat{h}_{\mathcal{M}}\circ \hat{g}_{\mathcal{M}}\right)}_{J_{11}}+\underbrace{\hat{r}\left(\hat{h}_{\mathcal{M}}\circ \hat{g}_{\mathcal{M}}\right)-\hat{r}\left(h^{\prime}_{\mathcal{M}}\circ \hat{g}_{\mathcal{M}}\right)}_{J_{12}}\\
     &+\underbrace{\hat{r}\left(h^{\prime}_{\mathcal{M}}\circ \hat{g}_{\mathcal{M}}\right)- r\left(h^{\prime}_{\mathcal{M}}\circ \hat{g}_{\mathcal{M}}\right)}_{J_{13}}\\
 \end{align}
 Clearly, $J_{12}\leq 0$ since $\hat{h}_{\mathcal{M}}$ is the minimizer of the empirical risk over $\mathcal{D}$ with the representation $\hat{g}_{\mathcal{M}}$. And $J_{11}+J_{13}\leq 2\sup_{h\in\mathcal{H}, g_{\mathcal{M}}\in\mathcal{G}_{\mathcal{M}}}\left|r(h\circ g_{\mathcal{M}})-\hat{r}(h\circ g_{\mathcal{M}})\right|$.
 %Assumption: bounded loss: C
 $$
\begin{aligned}
    &\sup _{h\in\mathcal{H},g_{\mathcal{M}}\in\mathcal{G}_\mathcal{M}}\left|\hat{r}\left(h\circ g_{\mathcal{M}}\right)-r\left(h\circ g_{\mathcal{M}}\right)\right|\\
    &= \sup _{h\in\mathcal{H},g_{\mathcal{M}}\in\mathcal{G}_{\mathcal{M}}}\left|\frac{1}{m}\sum_{i=1}^{m}\ell\left(h\circ g_{\mathcal{M}}(\mathbf{x}_{i}),y_i\right)-\mathbb{E}_{(\mathbf{x}^{\prime},y^{\prime})\sim \mathcal{D}}\left[\ell\left(h\circ g_{\mathcal{M}}(\mathbf{x}^{\prime}),y^{\prime}\right)\right]\right|
\end{aligned}
$$
% $$
%\begin{aligned}
%    &\sup _{h\in\mathcal{H},g_{\mathcal{M}}\in\mathcal{G}_\mathcal{M}}\left|\hat{r}\left(h\circ g_{\mathcal{M}}\right)-r\left(h\circ g_{\mathcal{M}}\right)\right|\\
%    &= \sup _{h\in\mathcal{H},g_{\mathcal{M}}\in\mathcal{G}_{\mathcal{M}}}\left|\frac{1}{m}\sum_{i=1}^{m}\ell\left(h\circ g_{\mathcal{M}},(\mathbf{x}_{i},y_i)\right)-\mathbb{E}_{(\mathbf{x}_{i}^{\prime},y_{i}^{\prime})\sim \mathcal{D}}\left[\frac{1}{m}\sum_{i=1}^{m}\ell\left(h\circ g_{\mathcal{M}},(\mathbf{x}^{\prime}_{i},y^{\prime}_i)\right)\right]\right|
%\end{aligned}
%$$
Since $\ell$ is bounded by a constant $C$, we have $0 \leq \ell\left(h\circ g_{\mathcal{M}}(\mathbf{x}), y\right) \leq C$ for any $(\mathbf{x}, y)$. As one pair $(\mathbf{x}_{i},y_{i})$ changes, the above equation cannot change by at most $\frac{2C}{m}$. Applying McDiarmid's\cite{crammer2008learning} inequality, we obtain that with probability $1-\delta/2$:
\begin{align}
&\sup _{h\in\mathcal{H},g_{\mathcal{M}}\in\mathcal{G}_{\mathcal{M}}}\left|\hat{r}\left(h\circ g_{\mathcal{M}}\right)-r\left(h\circ g_{\mathcal{M}}\right)\right|\\
\leq& \mathbb{E}_{(\mathbf{x}_{i},y_{i})\sim \mathcal{D}} \sup _{h\in\mathcal{H},g_{\mathcal{M}}\in\mathcal{G}_{\mathcal{M}}}\left|\frac{1}{m}\sum_{i=1}^{m}\ell\left(h\circ g_{\mathcal{M}}(\mathbf{x}_{i}),y_i\right)-\mathbb{E}_{(\mathbf{x}^{\prime},y^{\prime})\sim \mathcal{D}}\left[\ell\left(h\circ g_{\mathcal{M}}(\mathbf{x}^{\prime}),y^{\prime}\right)\right]\right|\label{rad}\\
&+ C\sqrt{\frac{2 \ln (2 / \delta)}{m}}
\end{align}
%\begin{align}
%&\sup _{h\in\mathcal{H},g_{\mathcal{M}}\in\mathcal{G}_{\mathcal{M}}}\left|\hat{r}\left(h\circ g_{\mathcal{M}}\right)-r\left(h\circ g_{\mathcal{M}}\right)\right|\\
%\leq& \mathbb{E}_{(\mathbf{x}_{i},y_{i})\sim \mathcal{D}} \sup _{h\in\mathcal{H},g_{\mathcal{M}}\in\mathcal{G}_{\mathcal{M}}}\left|\frac{1}{m}\sum_{i=1}^{m}\ell\left(h\circ g_{\mathcal{M}},(\mathbf{x}_{i},y_i)\right)-\mathbb{E}_{(\mathbf{x}_{i}^{\prime},y_{i}^{\prime})\sim \mathcal{D}}\left[\frac{1}{m}\sum_{i=1}^{m}\ell\left(h\circ g_{\mathcal{M}},(\mathbf{x}^{\prime}_{i},y^{\prime}_i)\right)\right]\right|\\
%&+ C\sqrt{\frac{2 \ln (2 / \delta)}{m}}
%\end{align}

To proceed the proof, we introduce a popular result of Rademacher complexity in the following lemma\cite{bartlett2002rademacher}:
\begin{lemma} \label{lemma}
Let $U, \left\{U_{i}\right\}_{i=1}^{m}$ be i.i.d. random variables taking values in some space $\mathcal{U}$ and $\mathcal{F} \subseteq[a, b]^{\mathcal{U}}$ is a set of bounded functions. We have
\begin{align}
    \mathbb{E}\left[\sup _{f \in \mathcal{F}}\left(\mathbb{E}[f(U)]-\frac{1}{m} \sum_{i=1}^{m} f\left(U_{i}\right)\right)\right] \leq 2 \mathfrak{R}_{m}(\mathcal{F})
\end{align}
\end{lemma}
\begin{proofof}{lemma 1}
Denote $\left\{U^{\prime}_{i}\right\}_{i=1}^{m}$ be ghost examples of $\left\{U_{i}\right\}_{i=1}^{m}$, i.e.
 $U_{i}^{\prime}$ be independent of each other and  have the same distribution as $U_{i}$. Then we have,
\begin{align}
& \mathbb{E}\left[\sup _{f \in \mathcal{F}}\left(\mathbb{E}[f(U)]-\frac{1}{m} \sum_{i=1}^{m} f\left(U_{i}\right)\right)\right] \\
=& \mathbb{E}\left[\sup _{f \in \mathcal{F}}\left(\frac{1}{m} \sum_{i=1}^{m}\left(\mathbb{E}[f(U)]-f\left(U_{i}\right)\right)\right)\right]\\
\stackrel{(a)}{=}& \mathbb{E}\left[\sup _{f \in \mathcal{F}}\left(\frac{1}{m} \sum_{i=1}^{m} \mathbb{E}\left[f\left(U_{i}^{\prime}\right)-f\left(U_{i}\right) \mid \left\{U_{i}\right\}_{i=1}^{m}\right]\right)\right] \\
\leq & \mathbb{E}\left[\mathbb{E}\left[\sup _{f \in \mathcal{F}}\left(\frac{1}{m} \sum_{i=1}^{m}\left(f\left(U_{i}^{\prime}\right)-f\left(U_{i}\right)\right)\right) \mid \left\{U_{i}\right\}_{i=1}^{m}\right]\right] \\
\stackrel{(b)}{=}& \mathbb{E}\left[\sup _{f \in \mathcal{F}}\left(\frac{1}{m} \sum_{i=1}^{m}\left(f\left(U_{i}^{\prime}\right)-f\left(U_{i}\right)\right)\right)\right] \\
=& \mathbb{E}\left[\sup _{f \in \mathcal{F}}\left(\frac{1}{m} \sum_{i=1}^{m} \sigma_{i}\left(f\left(U_{i}^{\prime}\right)-f\left(U_{i}\right)\right)\right)\right] \\
\leq & \mathbb{E}\left[\sup _{f \in \mathcal{F}} \frac{1}{m} \sum_{i=1}^{m} \sigma_{i} f\left(U_{i}^{\prime}\right)\right]+\mathbb{E}\left[\sup _{f \in \mathcal{F}} \frac{1}{m} \sum_{i=1}^{m} \sigma_{i} f\left(U_{i}\right)\right] \\
\stackrel{(c)}{=}& 2 \mathfrak{R}_{m}(\mathcal{F}) .
\end{align}

where $\sigma_{1}, \ldots, \sigma_{m}$ is i.i.d. $\{\pm 1\}$ -valued random variables with $\mathbb{P}\left(\sigma_{i}=+1\right)=\mathbb{P}\left(\sigma_{i}=-1\right)=1 / 2$. $(a)$ $(b)$ are obtained by the tower property of conditional expectation;  $(c)$ follows from the definition of Rademacher complexity of $\mathcal{F}$. 
\end{proofof}
Consider the function class:
$$\ell_{\mathcal{H}\circ\mathcal{G}_{\mathcal{M}}}:=\{(\mathbf{x}, y) \mapsto \ell\left(h\circ g_{\mathcal{M}}(\mathbf{x}),y\right)\mid h \in \mathcal{H}, g_{\mathcal{M}}\in \mathcal{G}_{\mathcal{M}}\}$$
let $\mathcal{F} = \ell_{\mathcal{H}\circ\mathcal{G}_{\mathcal{M}}}$ in lemma~\ref{lemma}, then we have equation $(\ref{rad})$ can be upper bound by $2\mathfrak{R}_{m}(\ell_{\mathcal{H}\circ\mathcal{G}_{\mathcal{M}}})$.
To directly work  with the hypothesis function class, we need to decompose the Rademacher term which consists of the loss function classes. We center the function $\ell^{\prime} (h\circ g_{\mathcal{M}}(\mathbf{x}),y)=\ell (h\circ g_{\mathcal{M}}(\mathbf{x}),y)-\ell (\mathbf{0},y)$. The constant-shift property of Rademacher averages\cite{bartlett2002rademacher} indicates that
$$
\mathfrak{R}_{m}(\ell_{\mathcal{H}\circ\mathcal{G}_{\mathcal{M}}})\leq \mathfrak{R}_{m}(\ell^{\prime}_{\mathcal{H}\circ\mathcal{G}_{\mathcal{M}}})+\frac{C}{\sqrt{m}}
$$
Since $\ell^{\prime}$ is Lipschitz in its first coordinate with constant $L$ and $\ell^{\prime} (h\circ g_{\mathcal{M}}(\mathbf{0}),y)=0$,  applying the contraction principle\cite{bartlett2002rademacher}, we have:
\begin{align*}
    \mathfrak{R}_{m}(\ell^{\prime}_{\mathcal{H}\circ\mathcal{G}_{\mathcal{M}}})\leq   2 L \mathfrak{R}_{m}( \mathcal{H}\circ{G}_{\mathcal{M}})
\end{align*}
Combining the above discussion, we obtain:
$$
J_1\leq 8L\mathfrak{R}_{m}( \mathcal{H}\circ{G}_{\mathcal{M}})+\frac{4C}{\sqrt{m}}+2C\sqrt{\frac{2 \ln (2 / \delta)}{m}}
$$
For $J_2$, by the definition of $h^{\prime}_{\mathcal{M}}$:
\begin{align}
    J_2&= \inf_{h_{\mathcal{M}}\in \mathcal{H}}\left[ r\left(h_{\mathcal{M}}\circ \hat{g}_{\mathcal{M}}\right)- r\left(\hat{h}_{\mathcal{N}}\circ \hat{g}_{\mathcal{N}}\right)\right]  \\
    &\leq \sup_{h_{\mathcal{N}}\in \mathcal{H} }\inf_{h_{\mathcal{M}}\in \mathcal{H} }\left[ r\left(h_{\mathcal{M}}\circ \hat{g}_{\mathcal{M}}\right)- r\left(h_{\mathcal{N}}\circ \hat{g}_{\mathcal{N}}\right)\right] \\
    &=\inf_{h_{\mathcal{M}}\in \mathcal{H}}\left[ r\left(h_{\mathcal{M}}\circ \hat{g}_{\mathcal{M}}\right)-r(h^{*}\circ g^{*})\right]-\inf_{h_{\mathcal{N}}\in \mathcal{H}}\left[ r\left(h_{\mathcal{N}}\circ \hat{g}_{\mathcal{N}}\right)-r(h^{*}\circ g^{*})\right]\\
    &= \eta(\hat{g}_{\mathcal{M}})-\eta(\hat{g}_{\mathcal{N}})\\
    &=\gamma_{\mathcal{S}}(\mathcal{M},\mathcal{N})
\end{align}
Finally,
$$
r\left(\hat{h}_{\mathcal{M}}\circ \hat{g}_{\mathcal{M}}\right)- r\left(\hat{h}_{\mathcal{N}}\circ \hat{g}_{\mathcal{N}}\right)\leq \gamma_{\mathcal{S}}(\mathcal{M},\mathcal{N})+8L\mathfrak{R}_{m}( \mathcal{H}\circ\mathcal{G}_{\mathcal{M}})+\frac{4C}{\sqrt{m}}+2C\sqrt{\frac{2 \ln (2 / \delta)}{m}}
$$
with probability $1-\frac{\delta}{2}$.
\end{proof}
\subsection{Proof of Theorem~\ref{thm-latent}}
\begin{proof}
Let $\tilde{h}_{\mathcal{M}}$ denote the minimizer of the population risk over $\mathcal{D}$ with the representation $\hat{g}_{\mathcal{M}}$, then we have:
%following the same analysis of the proof of Theorem 1, we have
\begin{align}
    &\eta(\hat{g}_{\mathcal{M}})\\
    &=r(\tilde{h}_{\mathcal{M}}\circ\hat{g}_{\mathcal{M}})-r(h^{*}\circ g^{*})\\
    &\leq \underbrace{r(\hat{h}_{\mathcal{M}}\circ\hat{g}_{\mathcal{M}})-\hat{r}(\hat{h}_{\mathcal{M}}\circ\hat{g}_{\mathcal{M}})}_{J_1}+\underbrace{\hat{r}(\hat{h}_{\mathcal{M}}\circ\hat{g}_{\mathcal{M}})-\hat{r}(h^{*}\circ g^{*})}_{J_2}+\underbrace{\hat{r}(h^{*}\circ g^{*})- r(h^{*}\circ g^{*})}_{J_3}
\end{align}
$J_2$ is the centering empirical risk. Following the similar analysis in Theorem~\ref{thm:mn-modality}, we obtain:
\begin{align}
    J_1+J_3&\leq \sup_{h\in\mathcal{H}, g_{\mathcal{M}}\in\mathcal{G}_{\mathcal{M}}}\left|r(h\circ g_{\mathcal{M}})-\hat{r}(h\circ g_{\mathcal{M}})\right|+\sup_{h\in\mathcal{H}, g\in\mathcal{G}}\left|r(h\circ g)-\hat{r}(h\circ g)\right|\\
    &\leq 4L\mathfrak{R}_{m}( \mathcal{H}\circ\mathcal{G}_{\mathcal{M}})+4L\mathfrak{R}_{m}( \mathcal{H}\circ\mathcal{G})+\frac{4C}{\sqrt{m}}+2C\sqrt{\frac{2 \ln (2 / \delta)}{m}}
\end{align}
with probability $1-\delta$. Combining the above discussion yields the result:
\begin{align}
     &\eta(\hat{g}_{\mathcal{M}})\leq\\
     &4L\mathfrak{R}_{m}( \mathcal{H}\circ\mathcal{G}_{\mathcal{M}})+4L\mathfrak{R}_{m}( \mathcal{H}\circ\mathcal{G})+\frac{4C}{\sqrt{m}}+2C\sqrt{\frac{2 \ln (2 / \delta)}{m}}+\hat{L}(\hat{h}_{\mathcal{M}}\circ\hat{g}_{\mathcal{M}},\mathcal{S})
\end{align}
\end{proof}
\subsection{Proof of Proposition 1}
\begin{proof}
With the $l_2$ loss, we have
$$
\mathbb{E}_{\mathbf{x}, y \sim h^{\star} \circ g^{\star}(\mathbf{x})}\{\ell(h \circ g(\mathbf{x}), y)-\ell(h^{\star} \circ g^{\star}(\mathbf{x}), y)\}=\mathbb{E}_{\mathbf{x}}\left[\left|\ \boldsymbol{\beta}^{\top}\mathbf{A}^{\top}\mathbf{x}-{\boldsymbol{\beta}^{\star}}^{\top}{\mathbf{A}^{\star}}^{\top}\mathbf{x}\right|^{2}\right]
$$
Define the covariance matrix\cite{tripuraneni2020theory} for two linear projections $\mathbf{A}$, $\mathbf{A}^{\prime}$ as follows:
\begin{align}
&\Gamma(\mathbf{A},\mathbf{A}^{\prime})
=\mathbb{E}_{\mathbf{x}}\left[\begin{array}{cc}
 \mathbf{A}^{\top}\mathbf{x}\left(\mathbf{A}^{\top}\mathbf{x}\right)^{\top}& \mathbf{A}^{\top}\mathbf{x}\left({\mathbf{A}^{\prime}}^{\top}\mathbf{x}\right)^{\top} \\
{\mathbf{A}^{\prime}}^{\top}\mathbf{x}\left(\mathbf{A}^{\top}\mathbf{x}\right)^{\top} & {\mathbf{A}^{\prime}}^{\top}\mathbf{x}\left({\mathbf{A}^{\prime}}^{\top}\mathbf{x}\right)^{\top}
\end{array}\right]\nonumber\\&=\left[\begin{array}{cc}
\mathbf{A}^{\top}\Sigma\mathbf{A} &  \mathbf{A}^{\top}\Sigma\mathbf{A}^{\prime} \\
 {\mathbf{A}^{\prime}}^{\top}\Sigma\mathbf{A}&  {\mathbf{A}^{\prime}}^{\top}\Sigma\mathbf{A}^{\prime}
\end{array}\right] =\left[\begin{array}{cc}
  \Gamma_{11}(\mathbf{A},\mathbf{A}^{\star})  & \Gamma_{12}(\mathbf{A},\mathbf{A}^{\star})  \\
   \Gamma_{21}(\mathbf{A},\mathbf{A}^{\star})  & \Gamma_{22}(\mathbf{A},\mathbf{A}^{\star})
\end{array}\right]
\end{align}
where $\Sigma$ denotes the covariance matrix of the distribution $\mathbb{P}_{\mathbf{x}}$. Then the \textit{latent representation quality} of $\mathbf{A}$ becomes:
\begin{align}
    \eta(\mathbf{A}) &= \inf_{\boldsymbol{\beta}:\|\boldsymbol{\beta}\|\leq C_b} \mathbb{E}_{\mathbf{x}}\left[\left|\ \boldsymbol{\beta}^{\top}\mathbf{A}^{\top}\mathbf{x}-{\boldsymbol{\beta}^{\star}}^{\top}{\mathbf{A}^{\star}}^{\top}\mathbf{x}\right|^{2}\right]\\
    &=\inf_{\boldsymbol{\beta}:\|\boldsymbol{\beta}\|\leq C_b}\left[\boldsymbol{\beta},-\boldsymbol{\beta}^{\star}\right]\Gamma(\mathbf{A},\mathbf{A}^{\star})\left[\boldsymbol{\beta},-\boldsymbol{\beta}^{\star}\right]^{\top}\label{mini}
\end{align}
    For sufficiently large $C_b$, the constrained minimizer of $(\ref{mini})$ is equivalent to the unconstrained minimizer. Following the standard discussion of the quadratic convex optimization \cite{boyd2004convex}, if $\Gamma_{11}(\mathbf{A},\mathbf{A}^{\star})\succ 0$ and $\det \Gamma_{11}(\mathbf{A},\mathbf{A}^{\star})\not=0$, the solution of the above minimization problem is $\boldsymbol{\beta} = \Gamma_{11}(\mathbf{A},\mathbf{A}^{\star})^{-1}\Gamma_{12}(\mathbf{A},\mathbf{A}^{\star})\boldsymbol{\beta}^{\star}$, and \begin{align}
        \eta(\mathbf{A}) = \boldsymbol{\beta}^{\star}\Gamma_{sch}(\mathbf{A},\mathbf{A}^{\star}){\boldsymbol{\beta}^{\star}}^{\top}
    \end{align}
    where $\Gamma_{sch}(\mathbf{A},\mathbf{A}^{\star})$ is the Schur complement of $\Gamma(\mathbf{A},\mathbf{A}^{\star})$, defined as:
    \begin{align}
        &\Gamma_{sch}(\mathbf{A},\mathbf{A}^{\star})\\ &= \Gamma_{22}(\mathbf{A},\mathbf{A}^{\star})-\Gamma_{21}(\mathbf{A},\mathbf{A}^{\star})\Gamma_{11}(\mathbf{A},\mathbf{A}^{\star})^{-1}\Gamma_{12}(\mathbf{A},\mathbf{A}^{\star})
    \end{align}
 Under the orthogonal assumption, $\hat{\mathbf{A}}_{\mathcal{M}}$ is  nonsingular. Notice that  $\hat{\mathbf{A}}_{\mathcal{N}}$ cannot be orthonormal in our settings. And $\sum$ is also invertible. Therefore, the Schur complement of  $\Gamma(\hat{\mathbf{A}}_{\mathcal{M}},\mathbf{A}^{\star})$ exists, 
\begin{align}
    \Gamma_{sch}(\hat{\mathbf{A}}_{\mathcal{M}},\mathbf{A}^{\star}) = {\mathbf{A}^{\star}}^{\top}\Sigma\mathbf{A}^{\star}-\left({\mathbf{A}^{\star}}^{\top}\Sigma\hat{\mathbf{A}}_{\mathcal{M}}\right)\left({\hat{\mathbf{A}}_{\mathcal{M}}}^{\top}\Sigma\hat{\mathbf{A}}_{\mathcal{M}}\right)^{-1}\left({\hat{\mathbf{A}}_{\mathcal{M}}}^{\top}\Sigma\mathbf{A}^{\star}\right)=\mathbf{0}
\end{align}
 Hence, $\eta(\hat{\mathbf{A}}_{\mathcal{M}})=0$.
 Given the above discussion, we obtain:
 \begin{align}
     \gamma_{\mathcal{S}}(\mathcal{M},\mathcal{N})&=\eta(\hat{\mathbf{A}}_{\mathcal{M}})-\eta(\hat{\mathbf{A}}_{\mathcal{N}})\\
     &= 0-\inf_{\boldsymbol{\beta}:\|\boldsymbol{\beta}\|\leq C_b} \mathbb{E}_{\mathbf{x}}\left[\left|\ \boldsymbol{\beta}^{\top}\hat{\mathbf{A}}_{\mathcal{N}}^{\top}\mathbf{x}-{\boldsymbol{\beta}^{\star}}^{\top}{\mathbf{A}^{\star}}^{\top}\mathbf{x}\right|^{2}\right]\leq 0
 \end{align}
\end{proof}
\section{The Composite Framework in Applications}\label{composite framework}
As we stated in Section~\ref{sec3}, our model well captures the essence of lots of existing multi-modal methods~\cite{baltruvsaitis2018multimodal,fayek2020large, feichtenhofer2016convolutional, wang2020deep,wang2020makes,kalfaoglu2020late}. Below, we explicitly discuss how these methods fit well into our general model, by providing the corresponding function class $\mathcal{G}$ under each method. 
\paragraph{Audiovisual fusion for sound recognition~\cite{fayek2020large}:} The audio and visual models map the respective inputs to segment-level representations, which are then used to obtain single-modal predictions, $\mathbf{h}_a$ and $\mathbf{h}_v$, respectively. The attention fusion function $n_{attn}$, ingests the single-modal predictions, $\mathbf{h}_a$ and $\mathbf{h}_v$, to produce weights for each modality, $\boldsymbol{\alpha}_{a}$ and $\boldsymbol{\alpha}_{v}$. The single-modal audio and visual predictions, $\mathbf{h}_{a}$ and $\mathbf{h}_{v}$, are mapped to $\tilde{\mathbf{h}}_{a}$ and  $\tilde{\mathbf{h}}_{v}$ via functions $n_{a}$ and $n_{v}$ respectively, and fused using the attention weights, $\boldsymbol{\alpha}_{a}$ and $\boldsymbol{\alpha}_{v}$. In summary, $g$ has the form: $$g = \tilde{\mathbf{h}}_{a v}=
\boldsymbol{\alpha}_a\odot \tilde{\mathbf{h}}_{a}+\boldsymbol{\alpha}_{v} \odot \tilde{\mathbf{h}}_{v}$$
\paragraph{Channel-Exchanging-Network~\cite{wang2020deep}:} A feature map will be replaced by that of other modalities at the same position, if its scaling factor is lower than a threshold. $g$ in this problem can be formulated as a multi-dimensional mapping $g:=(f_1,\cdots,f_M)$, where subnetwork $f_m(x)$ adopts the multi-modal data $x$ as input and fuses multi-modal information by channel exchanging.
\paragraph{Other Fusion Methods~\cite{feichtenhofer2016convolutional, wang2020makes,kalfaoglu2020late,baltruvsaitis2018multimodal}:} Methods in these works can be formulated into the form we mentioned in the example in Section~\ref{sec3}. Specifically, recall the example, $g$ has the form: $\varphi_{1} \oplus \varphi_{2} \oplus \cdots \oplus \varphi_{M}$, where $\oplus$ denotes a fusion operation, (e.g., averaging, concatenation, and self-attention), and $\varphi_{k}$ is a deep network which uses each modality data $x^{(k)}$ as input. Under these notations:
\begin{itemize}
    \item For the early-fusion BERT method in \cite{kalfaoglu2020late}, the temporal features are concatenated before the BERT layer and only a single BERT module is utilized. Here, the $\oplus$ is a concatenation function, and $g$ has the form $(\varphi_{1} ,\varphi_{2} )$. 
    \item \cite{wang2020makes,feichtenhofer2016convolutional}discussed different fusion methods by choosing $\oplus$. (i) Max fusion: the $\oplus$ is the maximum function and $g:=max \left\{\varphi_{1},\cdots, \varphi_{M}\right\}$; (ii) Sum fusion:  $g:=\sum \varphi_{m}$; (iii) averaging; (iv) self-attention and so on.
    \item The fusion section in the survey \cite{baltruvsaitis2018multimodal} provides many works which can be incorporated into our framework. 
\end{itemize}

\section{Discussions on Training Setting}
\label{appendix:training_setting}
Existing works on multi-modal training demonstrates that naively fusing different modalities results insufficient representation learning of each modality~\cite{wang2020makes, du2021improving}. In our experiments, we train our multi-modal model using two methods: (1), naively end-to-end late-fusion training; (2), firstly train the uni-modal models and train a multi-modal classifier over the uni-modal encoders. As shown in Table~\ref{tab:ienum} and Table~\ref{tab:ienum_stable}, naively end-to-end training is unstable, affecting the representation learning of each modality, while fine-tuning a multi-modal classifier over trained uni-modal encoders is more stable and the results are more consistent with our theory. 
Noting that we use the late-fusion framework here, similar to ~\cite{wang2020makes, du2021improving}.

\begin{table}[h]
   \centering
\begin{tabular}{cccccc}
 \toprule
  \cr \multicolumn{1}{c}{Modalities}   & \multicolumn{5}{c}{ Test Acc (Ratio of Sample Size)} \cr
 \cmidrule(lr){2-6}
 & $10^{-4}$ & $10^{-3}$&$10^{-2}$&$10^{-1}$& 1\cr
\midrule
 T   & 23.66$\pm$1.28&29.08$\pm$3.34& 45.63$\pm$0.29&48.30$\pm$1.31& 49.93$\pm$0.57\cr
 TA &  22.74$\pm$1.86&35.14$\pm$0.38 &49.15$\pm$0.43&50.61$\pm$0.28& 51.78$\pm$0.08\cr
 TV &  23.64$\pm$0.07&36.64$\pm$1.79 &46.91$\pm$0.68& 48.96$\pm$0.47&53.24$\pm$0.35\cr
 TVA & \textbf{25.40$\pm$1.06}&\textbf{40.87$\pm$2.47}& \textbf{50.67$\pm$0.63}&\textbf{52.54$\pm$0.60}&\textbf{54.55$\pm$0.29}\cr
 \bottomrule
\end{tabular}
\caption{Latent representation quality vs. The number of the sample size on IEMOCAP. In this table, we fristly train the uni-modal models and train a multi-modal classifier over the uni-modal encoders to get multi-modal results.}
\label{tab:ienum_stable}
\end{table}

\end{document}